\documentclass[letterpaper]{article}

\usepackage{natbib,alifeconf}  
\usepackage{url}

\title{Being curious about the answers to questions: novelty search with learned attention}
\author{Nicholas Guttenberg$^{1,2}$, Martin Biehl$^{1}$, Nathaniel Virgo$^{2}$, Ryota Kanai$^{1}$ \\
\mbox{}\\
$^1$ Araya Inc, Tokyo, Japan
$^2$ Earth-life Science Institute, Tokyo, Japan}

\begin{document}
\maketitle

\setlength{\parskip}{0pt}

\begin{abstract}
We investigate the use of attentional neural network layers in order to learn a `behavior characterization' which can be used to drive novelty search and curiosity-based policies. The space is structured towards answering a particular distribution of questions, which are used in a supervised way to train the attentional neural network. We find that in a 2d exploration task, the structure of the space successfully encodes local sensory-motor contingencies such that even a greedy local `do the most novel action' policy with no reinforcement learning or evolution can explore the space quickly. We also apply this to a high/low number guessing game task, and find that guessing according to the learned attention profile performs active inference and can discover the correct number more quickly than an exact but passive approach.
\end{abstract}

\section{Introduction}

There are now a number of approaches designed to drive exploration of unseen spaces. Intrinsically motivated curiosity algorithms drive reinforcement learning agents and agents which learn models alike into low-likelihood states \citep{ostrovski2017count, achiam2017surprise}, hard-to-predict states \citep{pathak2017curiosity}, or just directly try to maximize the information gained by the agent about its world \citep{friston2015active, houthooft2016vime, de2018curiosity}. In developmental robotics, goal babbling \citep{baranes2013active} drives an agent to map out its ego-motion space before committing to any particular reward. Similarly, there are methods in machine learning which use diversity-based metrics to learn sets of skills \citep{guttenberg2017learning, eysenbach2018diversity}. 

These methods share the aspect that they require signficant learning of the environmental dynamics, long-term planning, and other aspects of the world. At the other end of the spectrum, there are a number of heuristic approaches which can be successfully curious starting from scratch. Umbrella sampling methods \citep{torrie1977nonphysical} in computational chemistry work by simply driving simulations away from frequently visited states by modifying the energy, while non-equilibrium versions achieve similar results by resampling trajectories that head towards rare areas of the space \citep{dickson2010enhanced}. In evolutionary algorithms, the general concept of quality diversity \citep{pugh2016quality} and more specifically novelty search \citep{stanley2011} has been used in a similar way, explicitly modifying the fitness of solutions with respect to how different they are from other attempted solutions. Novelty search has also been applied to modify reward functions used in reinforcement learning \citep{conti2017improving}.

In the specific case of novelty search, there is a balance between the pure heuristic of exploring rare states and the learned approaches which in some sense try to collapse abstractly equivalent but microscopically different states into the same representation. In novelty search, this takes the form of the behavior characterization --- the way in which distinct agent behaviors or outcomes are embedded into a Euclidean space in order to assess their novelty. While much of the work uses hand-crafted behavior characterizations based on some knowledge of the relevant degrees of freedom of the task at hand, there has been work in formulating general characterizations which work across many tasks \citep{doncieux2013behavioral}, and in learning the characterizations with respect to specific criteria for the quality of exploration \citep{meyerson2016learning}.

In this paper, we consider a correspondence between recent ideas about attention in deep learning and the novelty search algorithm, both of which have embedding spaces at their core. In attentional neural networks \citep{vaswani2017attention}, the network takes as an input a collection of memories, and then learns to perform a lookup on that collection of memories in order to extract information that is relevant towards answering a particular question. The attentional network achieves this by first embedding both the memories and the `query' (which encodes information specifying the question) into a shared space where location in the space is informative about relevancy, and then uses pairwise comparison between the query and memories to determine which particular information from the memories to use. This is quite similar to how behavior characterization learns an embedding space that captures novelty. However, in the case of novelty, what is being represented is the potential to be relevant for any question rather than a specific one. Because of this, novelty search makes use of distance from the distribution of existing experiences rather than proximity to a specific query. If we use a distance metric for pairwise comparison in the attentional networks, the learned space should equally well be able to evaluate novelty, and as such we can bridge the two methods.


\section{Attentional Networks}

Traditional neural networks are composed of a series of matrix multiplications and nonlinearities, where the matrices correspond to parameters learned via gradient descent:

\begin{equation}
 \vec{y} = f( \textbf{M}_n f( \textbf{M}_{n-1} f( ... f( \textbf{M}_1 \vec{x} + \vec{b_1} ) ... ) + \vec{b}_{n-1} ) + \vec{b_n} )
\end{equation}

In this case, it is easy to compute the gradient of the parameter matrices (or `weights') $\textbf{M}$ with respect to some error function using only a backpropagated error signal that does not depend on the rest of the network. The recent availability of automatic differentiation has enabled rapid experimentation in a wide variety of alternate structures: extending to general higher-dimensional tensors, replacing matrices with convolution operations, using branching and merging patterns, factorizing the matrix multiplications, so-called `highway' or `residual' network structures which sequentially perturbatively modify the network's hidden state, and various forms of regularization.

A common pattern among this zoo of different network architectures is that as information from a particular input flows through the network, the relationship between the input and the intended output becomes potentially more and more complex, but also becomes more difficult to learn via gradient descent. The reason for this is that each time the input is multiplied by a parameter matrix in series, the various gradients are scaled by the eigenvalues of that matrix --- meaning that for very deep paths, the gradients tend to either diverge to infinity or converge to zero. Techniques such as orthogonal initialization \citep{saxe2013exact} help relax these constraints, allowing modern networks to take advantage of the nonlinearity offered by deep stacks, but in cases where information is unnecessarily propagated through many subsequent layers it can still pose difficulties in terms of training times. As such, there are trade-offs involved in simultaneously achieving a wide receptive field (in that the network can integrate large amounts of evidence towards some prediction or inference) while maintaining pathways through the network for information to flow which are not any longer than necessary.

This consideration makes random access memory an attractive model for network design. If the network could simply choose which information it needed to look at for each stage of the computation without needing to propagate that information through a large number of intervening variables, then one could minimize the amount of unnecessary scattering happening between the input and output. There is a general class of methods which implement this sort of random access via having each potential input be matched against a `query' generated by the current stage in the computation, such that the information brought into the network at that stage is a weighted sum across the inputs --- these are generally referred to as attention mechanisms. Such attention mechanisms have the property that the minimum path length to create a receptive field covering the entire input dataset is constant, irrespective of the dataset size \citep{vaswani2017attention}.

The general structure of these networks has each potential input (where each input is a vector $\vec{x}_i$) generate a key vector $\vec{k}_i$ summarizing the information it contains via linear transformation, while the current computational state $\vec{z}$ generates one or more query vectors $\vec{q}$ in the same space again via linear transformation. The saliency of an input $w_i$ is determined by comparing these key and query vectors --- often, via the exponential of the dot product of the vector. Then, the overall saliency pattern is normalized to sum to one:

\begin{equation}
w_i = \frac{\exp(\vec{k}_i \cdot \vec{q})}{\sum_j \exp(\vec{k}_j \cdot \vec{q})}
\label{DotProductAttention}
\end{equation}

where $\vec{k}_i = \textbf{M}_k \vec{x}_i$ and $\vec{q} = \textbf{M}_q \vec{z}$. The input to the network from this attentional lookup is then the weighted sum over the inputs:

\begin{equation}
\vec{X} = \sum_i w_i \vec{x}_i
\end{equation}

This formulation is fully differentiable, but can at the same time learn to attend to only a small subset of the total input. An interesting consequence of learning to solve problems using this type of attention model is that the saliency pattern $w$ weights inputs according to their relevance to the current computation, and so can be directly inspected or used for interpretation of how the network is solving a particular task.

In this paper, we exploit that property of $w$ to evaluate how relevant potential inputs to a network would be, and in turn use that pattern of projected relevance in order to direct the actions of an agent to search for relevant information in its environment. By making use of a variation in which the comparison between key and query is based not on a dot product, but rather on the distance between the vectors in a Euclidean space, we can bridge attentional neural networks with existing research done on behavior characterizations in novelty search.

\section{Method}

\begin{figure}
 \includegraphics[width=\columnwidth]{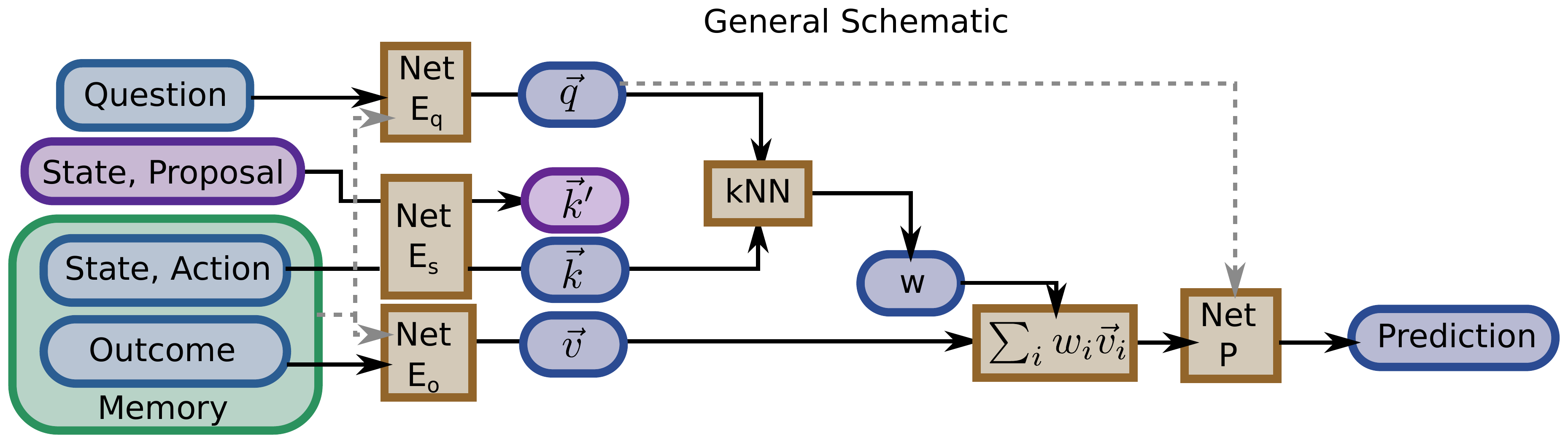}
 \caption{\label{Schematic}General schematic of an inference network which also learns a behavior characterization and can evaluate the saliency of a given proposed action to objectives such as novelty search or question-motivated curiosity. The dashed lines represent optional information flows that can be added without disrupting the ability to correctly embed action proposals.}
\end{figure}

If we wished to make an agent that could answer any one of a set of questions which might be posed to it (similar to ideas of multi-task behavior characterizations in  \cite{meyerson2016learning}), we could ask an agent to fill its memory with a set of points that is likely to contain a good match from any random query to which it might be exposed. This is then heuristically satisfied by finding new points to add to memory which are furthest from the set of existing points --- the same criteria which drives the novelty search algorithm. In this case however, the behavior characterization can be naturally learned directly from the set of questions the agent is trained to answer.

In novelty search, the weighting function used is essentially a nearest neighbor look-up in the embedding space: $1/k$ if a point is one of the k-nearest neighbors, and $0$ otherwise. In order to use this as part of a neural network, it is convenient to modify it into a differentiable form so that we can learn the embedding via backpropagation. To this end, we can consider an extension in the form of a `soft kNN' similar to the trick used in  \citep{pritzel2017neural, jain2018semiparametric} by proposing a weighting between the embeddings of the agent's memory of past experiences $\vec{x}_i$ which is used to generate a key vector $\vec{k}_i$ and value vector $\vec{v}_i$ via linear transformation and a query vector $\vec{q}$, given by:

\begin{equation}
w_i = \frac{\exp(-\alpha |\vec{k}_i - \vec{q}|)}{\sum_j \exp(-\alpha |\vec{k}_j - \vec{q}|) }
\end{equation}

where the output of this soft kNN module is:

\begin{equation}
\vec{y} = \sum_i w_i \vec{v}_i
\end{equation}

In order to tie this to inference on a particular question, we consider the general type of network structure shown in Fig.~\ref{Schematic}. Here, we have some kind of representation of the question and relevant context (current sensor state, etc), which we transform into the query vector $\vec{q}$ by way of an embedding network $E_q$. This transformation can optionally make use of information in the agent's memory (the dashed line) by way of an attentional mechanism. We also have a memory of past sensor states ($s$), actions ($a$), and outcomes ($o$) (which could be subsequent sensor states, a different set of sensors, etc). We use a network $E_s$ to embed the states and actions into a set of key vectors $\{\vec{k}\}$ representing the agent's memory of past states and actions (but not what happened next, e.g. the outcomes). We also separately embed the outcomes into a set of corresponding value vectors $\{\vec{v}\}$ (which can optionally be fused with other memory information, or even be fused with information about the question) using a network $E_o$. 

Weighting factors are calculated between $\vec{q}$ and the set $\{\vec{k}\}$, and then are used to perform a weighted average over the outcome embeddings. These (optionally fused with the query) are then used as inputs to a prediction network $P$, and the entire thing is trained end-to-end.

The reason for this structure is that it isolates things which the agent already knows or can directly control (current sensor state and future action) from things which depend on environmental factors (the outcome of taking an action). This means that if we propose some action $\tilde{a}$ from the current sensor state $\tilde{s}$ we can directly compute the address where that new generated memory will end up in the embedding space. That allows us to take as a behavioral policy things such as `choose the action which generates the most novel memory' or `choose an action which will be most relevant towards a specified question'. This means that by setting up the network this way, we can obtain curiousity-driven policies that do not require reinforcement learning or evolutionary search by greedily taking the most novel local action. As a result, these policies can be used directly by an agent dropped into a new environment and adapted to the new environment in an online fashion.

\section{Position-based Exploration Task}

We consider an environment comprised of navigating connected 2D floorplans composed of overlapping randomly generated rectangles of movable area, in which an agent is navigating by picking a direction and moving in that direction for a fixed distance. Collisions result in the agent reversing direction and continuing to travel for the remainder of its movement distance. The agent has two sensors: one which provides its current coordinates, and a second which indicates when it collides with an object or wall. For the inference question, we ask the agent to predict the probability that a given action will cause it to collide with a wall, meaning that the agent's task is essentially to thoroughly map out the boundaries and collidable surfaces of its environment.

The state is taken to be the current position, the action being the $x$ and $y$ components of the proposed direction of travel (normalized to a unit vector), and the observation $o$ to be whether or not the agent collided with a wall. Since the input sensors are so simple, we don't necessarily expect that much of a difference in the embedding space compared to a trivial behavior characterization where we just use everything. However, this simple case does test one somewhat non-trivial thing --- namely, the agent does not have any prior knowledge about its sensor-motor coupling. That is to say, even if the agent could identify a nearby physical location which it hasn't been to before, that is not a priori associated with the action which will actually take it to that location. As such, with this environment we are testing whether or not the agent can learn how to navigate to novel regions purely as a byproduct of learning to solve the particular inference task of figuring out if it will make contact with a wall.

For this task, the question takes the form of a $(s,a)$ pair (from this position, if you move in this direction, will you hit a wall?), so we use the same network for both $E_q$ and $E_s$. This network is composed of four hidden layers of size $256$ with rectified linear nonlinearities (ReLU), followed by a learned linear projection into a $24d$ embedding space. The network $E_o$ takes as input the full $(s,a,o)$ triplets and passes them through four hidden layers of size $256$ with ReLU nonlinearities, followed by linear projection into a $128d$ space. The soft kNN layer combines these ($\alpha = 20$), and finally the prediction network $P$ passes that result through four more hidden layers (size $256$, ReLU activations), with a final single-unit layer with sigmoid activation to output the predicted probability that the agent will collide with a wall.

The network is trained using the logistic loss $L = -y\log(p)-(1-y)\log(1-p)$ on $1000$ trajectories from different, randomly generated floorplans, where each trajectory is $1000$ steps long and is generated by taking random actions. During training, batches are composed by randomly selecting $50$ of these trajectories and choosing a random start point, such that the agent remembers the previous $300$ steps before the start point and must predict the collisions over the next $100$ steps. The agent is trained on $15000$ of these batches using the Adam optimizer \citep{kingma2014adam} with an initial learning rate of $4\times 10^{-4}$ that is reduced by a factor of $0.9$ whenever the test loss does not decrease for $10$ iterations. All of the networks are implemented and trained using PyTorch \citep{paszke2017automatic}. All code for the experiments in this paper is provided at \url{https://github.com/arayabrain/QuestionDrivenNovelty}.

\begin{figure}
\includegraphics[width=\columnwidth]{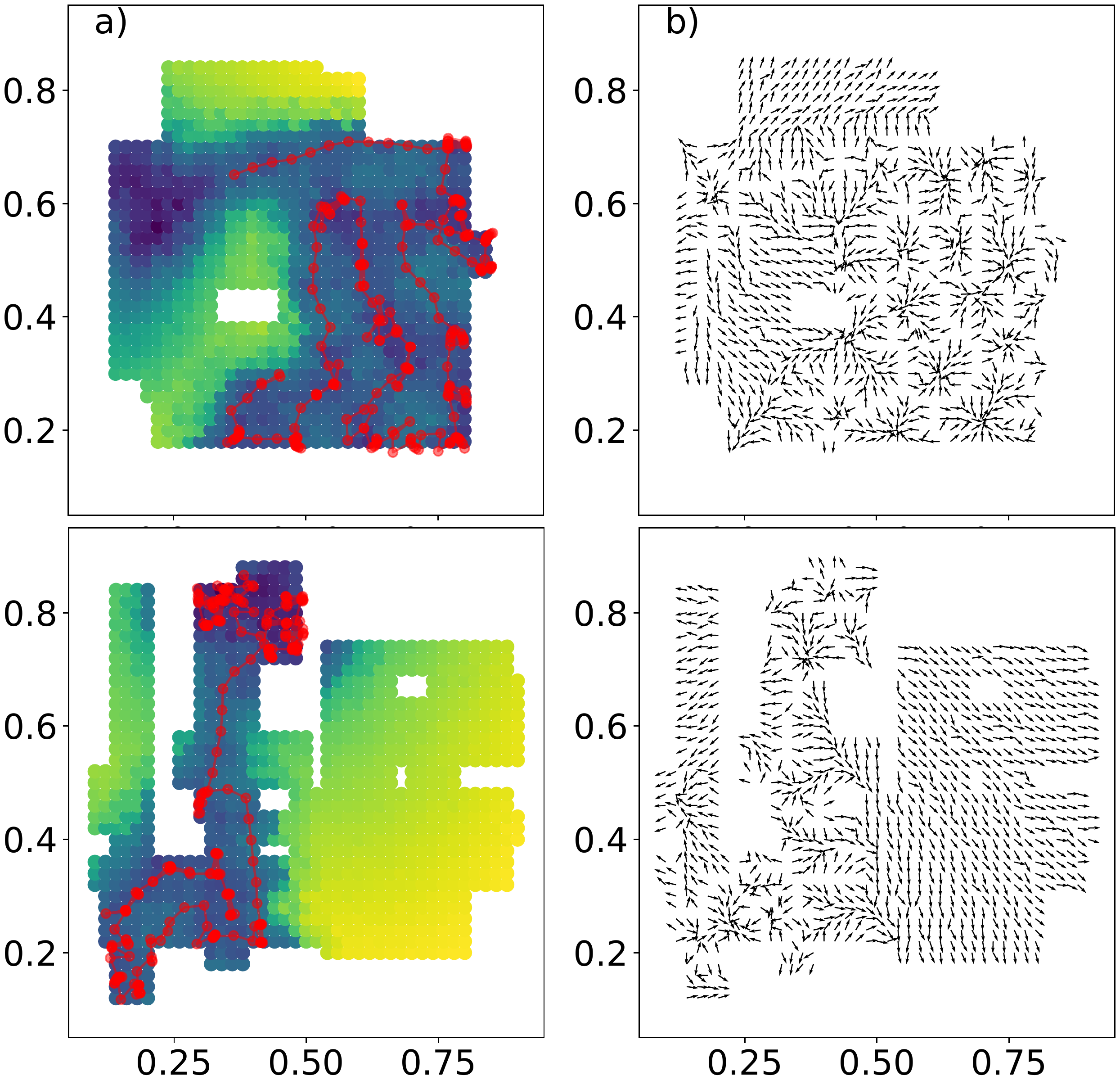}
\caption{ \label{Trajectory} a) Example floorplan and 300 steps of trajectory for the curious agent. The background color corresponds to the agent's current evaluation of novelty by position at the end of the trajectory, with yellow corresponding to high novelty and blue corresponding to low novelty. b) Vector field showing the action at each location which the agent would find most novel. }
\end{figure}

In order to `behave curiously', we seed the agent with $50$ steps of random actions to have an initial memory, and then for each subsequent step we propose $50$ random actions $\tilde{a}$ and choose the action which maps to a point in the embedding space $z^\prime$ which is furthest from its nearest point in memory. An example floorplan and trajectory over 300 subsequent steps is shown in Fig.~\ref{Trajectory}a. The background of this figure corresponds to a visualization of the model's assessments of the novelty at different locations, generated by creating a grid of points over the possible $(x,y)$ coordinates within the floorplan, proposing actions taken from each of those points, and taking the distance associated with the most `novel' action at each point. This shows that, predictably, the model considers points where it has never been to be novel compared to points which it has already visited. When we compare this to the direction of the most novel action from each point (Fig.~\ref{Trajectory}b), we find good agreement between the proposed actions and the direction of the local spatial gradient of the novelty --- meaning that the agent has apparently correctly learned the sensor-motor contingencies in order to navigate the space. At the same time (due to lack of feedback between the contents of memory and the embeddings), the novel actions are only locally informative, but do not correctly take into account long-range planning of paths through the space to reach high-novelty regions.

\begin{figure}
 \includegraphics[width=\columnwidth]{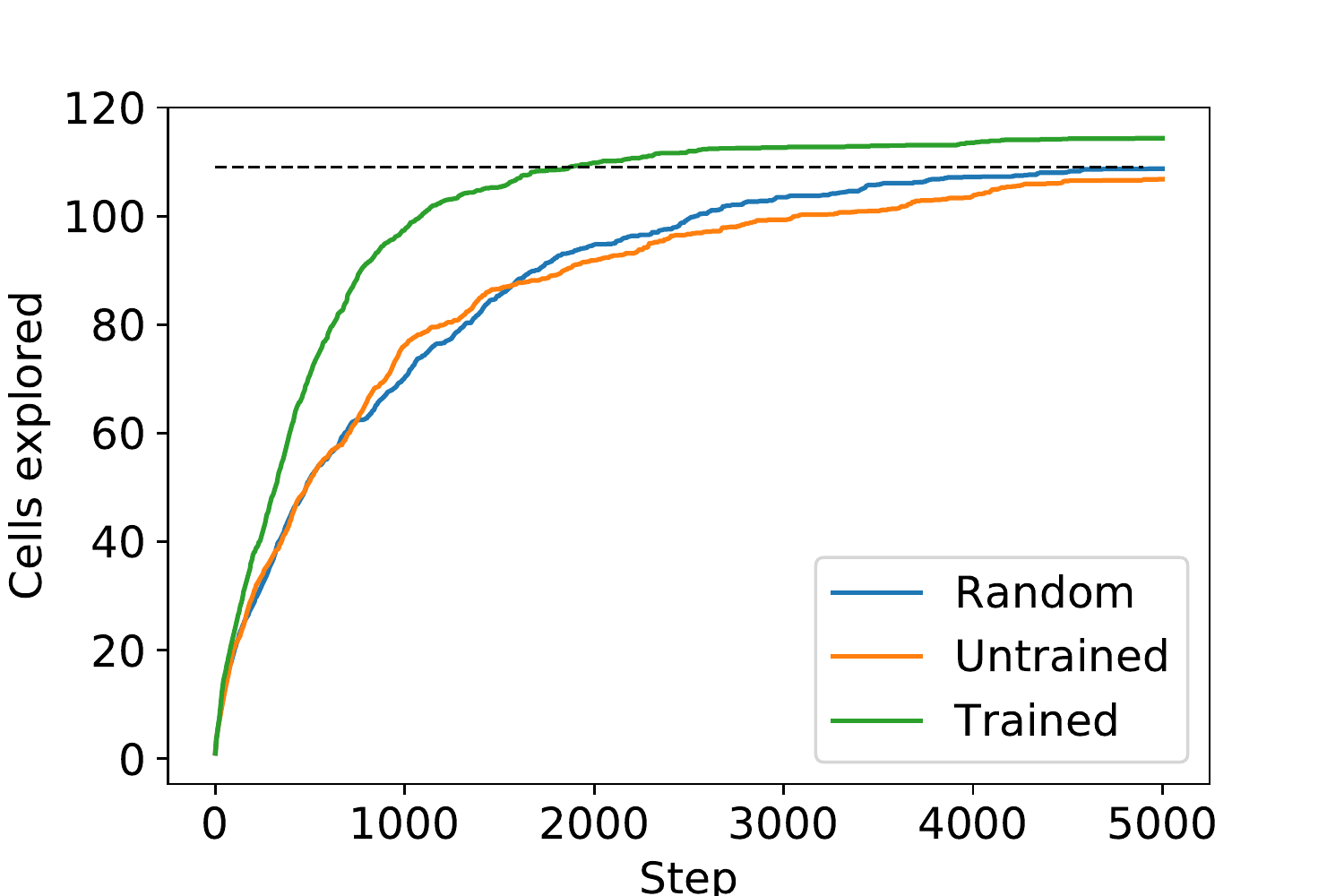}
 \caption{\label{Exploration} Exploration of new grid cells by the random action policy, the untrained network policy, and the trained network policy. These are curves are averaged over $15$ newly generated random floorplans. }
\end{figure}

Given that the agent seems to be able to pursue high novelty regions, does the learned behavior characterization provide reasonable direction as to efficient exploration? We compare the rates at which a random action policy, the untrained network, and the trained network explore the space by dividing the domain into a $16 \times 16$ grid and measuring over $15$ random (unseen) floorplans the average number of grid cells explored as a function of time. This is shown in Fig.~\ref{Exploration}. The trained network policy explores significantly faster than the random walk, roughly reaching the same level of exploration after $1900$ steps as the random action policy obtains in $5000$ steps.

\begin{figure}
\includegraphics[width=\columnwidth]{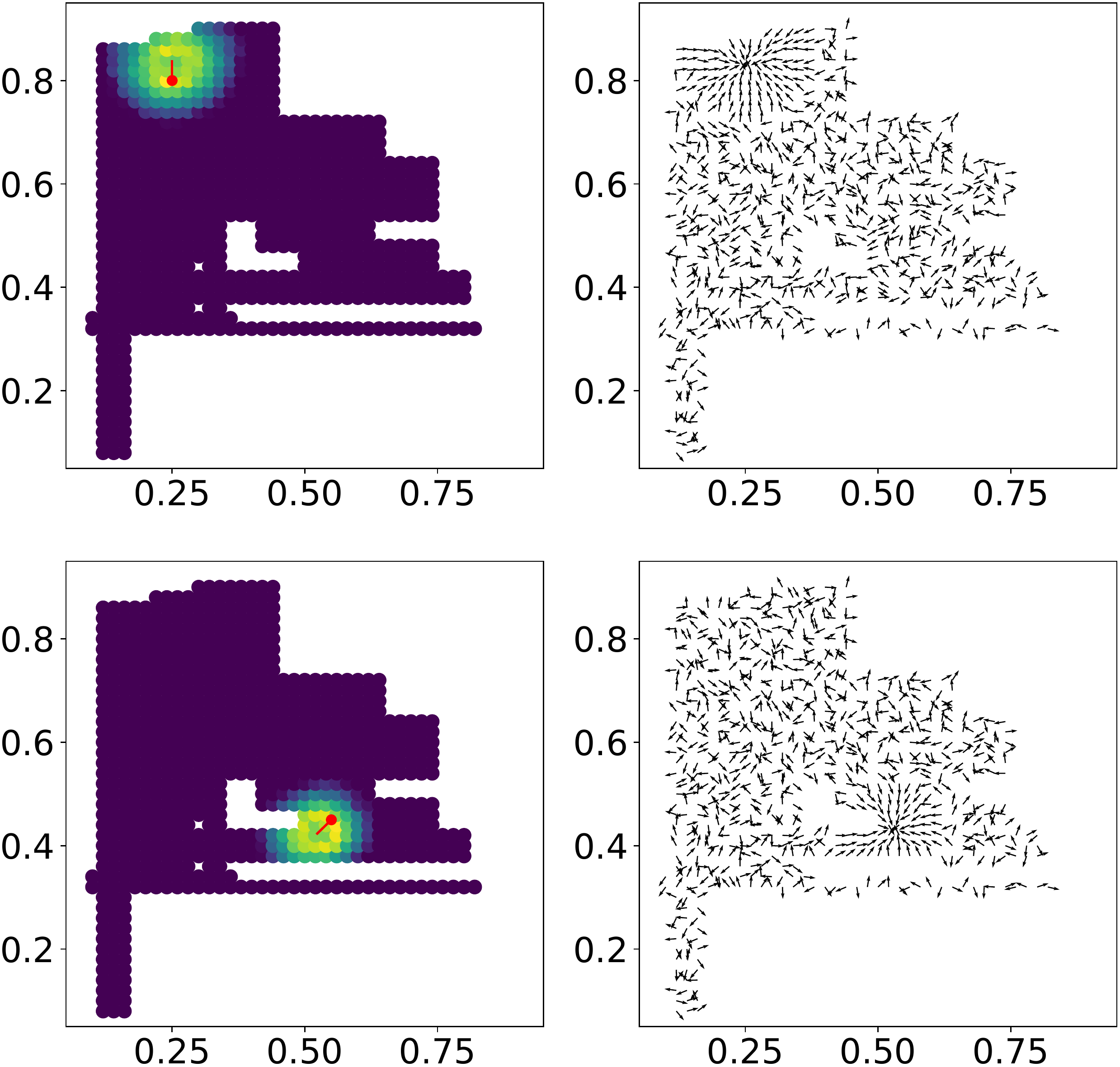}
\caption{ \label{QuestionSaliency} Plot of the saliency of different positions towards specific questions (will a collision occur in this direction from this point?), and the corresponding greedy saliency-maximizing policy. }
\end{figure}

When we ask the network to follow a novelty search policy, we are in essence asking it to take actions which can help it answer questions for which it does not yet have good supporting data. If we have a particular question in mind, such as whether there would be a wall at a particular point in space, we can directly visualize the actions and positions which the network would consider informative towards answering that question. We do this by projecting the question into the embedding space, and then visualizing the attention given to a virtual grid of state/action pairs. Visualizations of this are shown in Fig.~\ref{QuestionSaliency}. The saliency associated with a spatially localized query falls off on the order of one or two of the agent's steps. This means that while a single question can be used to locally guide the agent, it does not provide globally consistent navigational cues if the agent is currently far away from the region in which the question could be answered, though it could still be used as a reward signal for reinforcement learning or evolutionary search.

\section{Guessing game}

We now turn to a more abstract system, to take specific advantage of the fact that the embedding space for the agent is organized to help the agent be curious about specific questions rather than necessarily curious in general. For this, we look at a game in which the agent must guess a number between 1 and 256, and is told whether their guess is high, low, or correct. This game constitutes an active inference task, where the early questions should be arranged in a binary search in order to provide the evidence that the agent will need in order to infer the correct number rapidly. An agent which guesses randomly and then renders a prediction at the end will significantly underperform compared to binary search, as will even an agent which guesses uniformly from the current possible set of values.

\begin{figure}
 \includegraphics[width=\columnwidth]{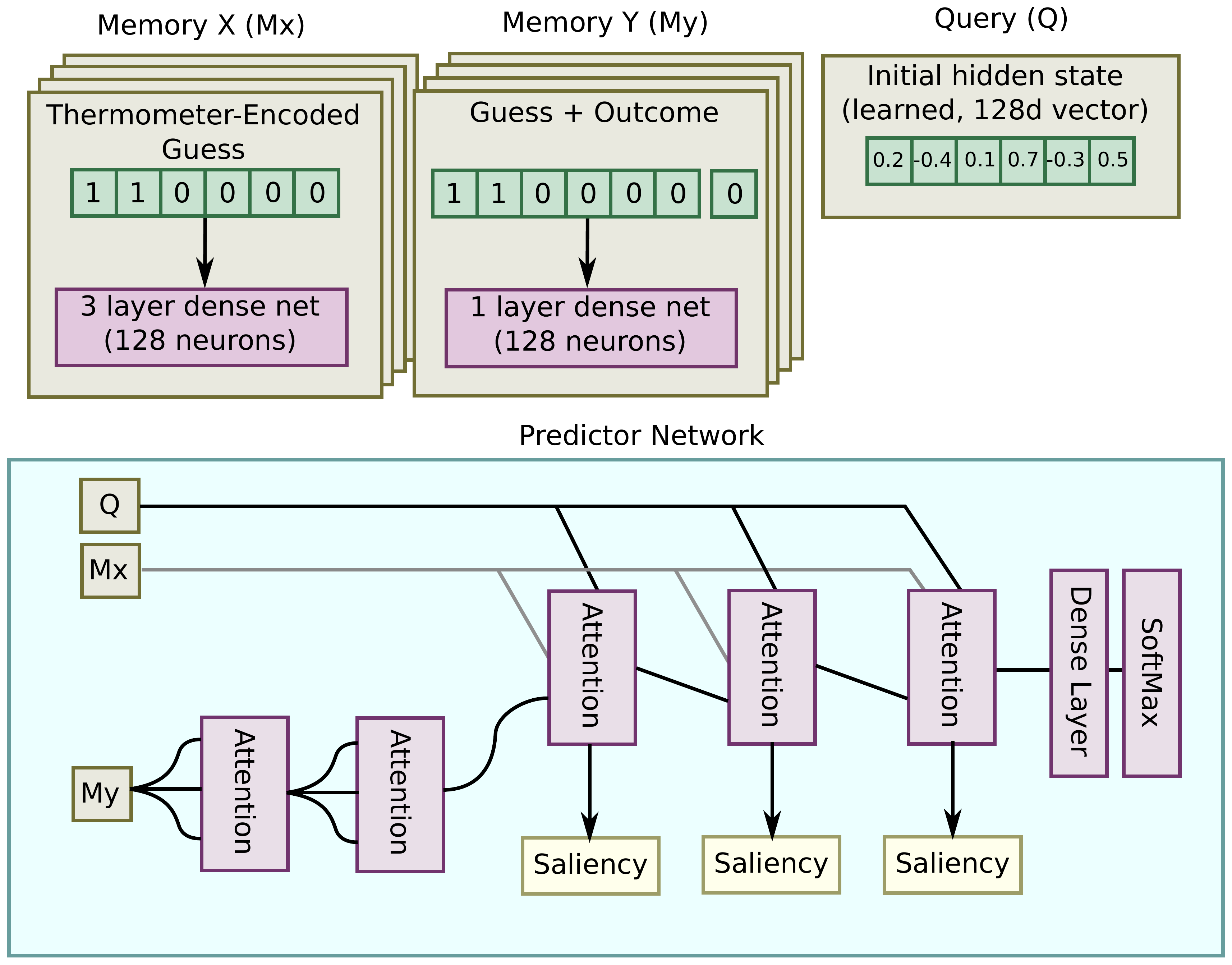}
 \caption{\label{GuessingArchitecture} Architecture of the guessing game network, using attention blocks as defined in Eq.~\ref{Block1} and Eq.~\ref{Block2}. Virtual queries for currently unexplored guesses are added to $M_x$, and saliency from the attentional layers is averaged in order to determine the most potentially informative guess to make next. }
\end{figure}

The network for this task is set up so that the internal process of paying attention to its own memory has a similar form to the guessing game structure. The network starts from a learned initial state vector, and has to update that state vector through a succession of three attentive lookups into its memory, indexed by the guess but not the outcome. As such, the network sort of plays a version of the guessing game internally in order to figure out its predictions and next guess, and the saliency profiles from those attentive lookups end up being good proxies for where the network would benefit from receiving more information.

Specifically, guesses are thermometer-encoded \citep{fiete2007neural} (with each guess $g$ being a $256$d vector whose first $g$ entries are set to one and the rest zero) and passed through three $128$ neuron hidden layers with ReLU activations --- this constitutes the $(s,a)$ embedding network $E_s$. One-hot encoded outcomes (high, low, equal) are stacked with the encoded guesses, passed through a single $128$ neuron hidden layer + ReLU, and then the outcomes cross-reference eachother using two attentive blocks constructed along the lines of the Transformer architecture \citep{vaswani2017attention} --- that is to say, using dot product attention as described in \ref{DotProductAttention} rather than Euclidean distance, combined with additional normalization and linear transformation steps. Each block has the form:

\begin{equation}
z_* \equiv \textrm{Normalize}( \textbf{z}_{in} + \textrm{Attn}(\textbf{z}_{in}, \textbf{z}_{in}) )
\label{Block1}
\end{equation}
\begin{equation}
z_{out} \equiv \textrm{Normalize}( \textbf{z}_* + \textbf{M}_2 \textrm{ReLU}(\textbf{M}_1 \textbf{z}_* + \vec{b}_1) + \vec{b}_2 )
\label{Block2}
\end{equation}

where $\textrm{Attn}(\textbf{x},\textbf{y})$ is an attentional lookup into the matrix of observations $\textbf{x}$ for each element of the matrix of queries $\textbf{y}$; $\textbf{M}_1$, $\textbf{M}_2$ are weight matrices with the same geometry as $z$, and $\vec{b}_1$ and $\vec{b}_2$ are biases (which are broadcast over the rows of $\textbf{z}$). The function $\textrm{Normalize}(\textbf{x})$ is the layer normalization \citep{ba2016layer} operation which subtracts the mean over features (columns of $\textbf{x}$) and divides by the standard deviation:

\begin{equation}
\textrm{Normalize}(\textbf{x})_{ij} = \frac{x_{ij} - \frac{1}{N} \sum_j x_{ij}}{\sqrt{\frac{1}{N} \sum_j x_{ij}^2 - (\frac{1}{N} \sum_j x_{ij})^2}}
\end{equation}

\begin{figure}
\includegraphics[width=\columnwidth]{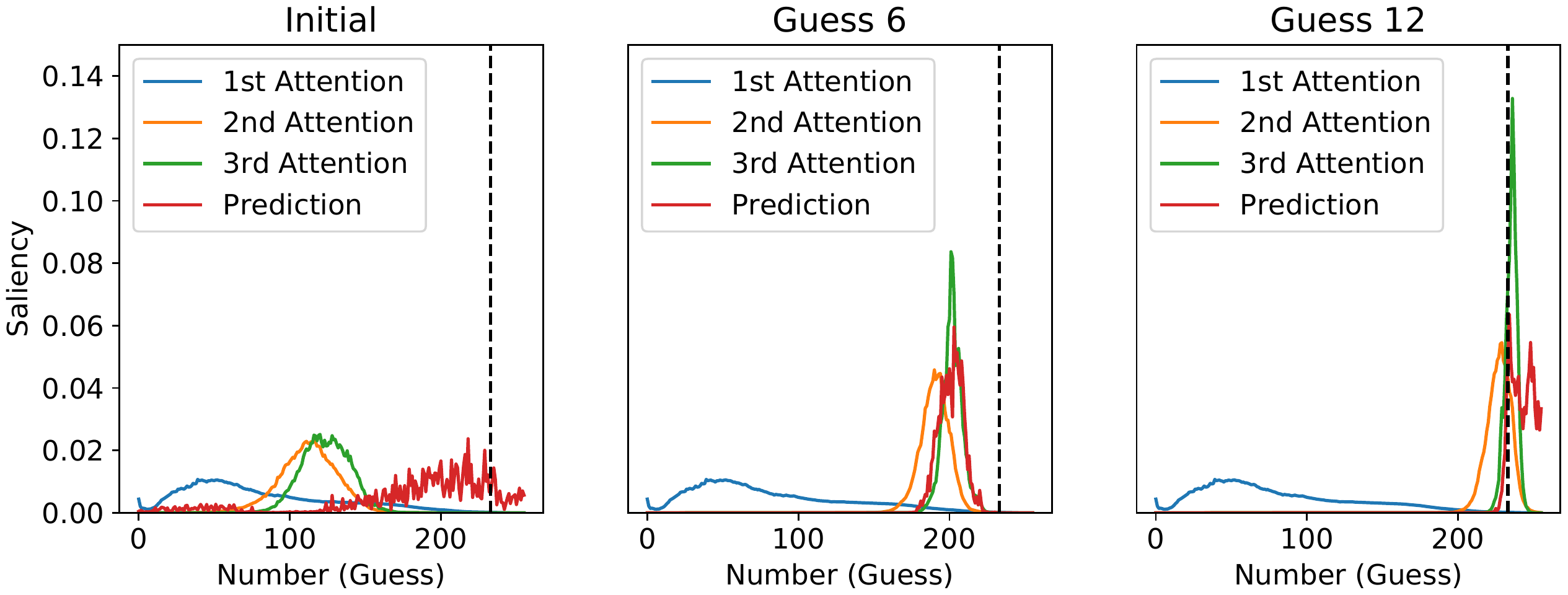}
 \caption{\label{Saliency} Saliency patterns from the trained guessing game network. The vertical dashed line shows the correct number, and the curves show the three query attention patterns after different numbers of guesses. Also visualized is the predicted probability distribution from the network.}
\end{figure}

Finally, the query encoder $E_q$ takes as input a learned initial query vector, which then drives three successive attentional blocks of the above form (with the last block replacing its input rather than adding to the input, in order to make sure all of the information used to generate the prediction comes from memory). The embedding spaces are all $8$d spaces, but different ones from $(s,a)$, $o$, and $q$ in each case. We look at the pattern of saliencies of these three passes from a trained network in Fig.~\ref{Saliency}. Finally, the output of the third attentional block is passed through two $128$ neuron hidden layers with ReLU activations followed by a softmax activation over $256$ values in order to render the probability distribution of the network's guess.

In order to encourage the agent to place saliency in future guesses that will be informative, we give the agent access to its future memories during training (but not during test). What this means is that the agent starts from a state that knows all about the game up to some move $t$, but then is issuing a query into a memory embedding space that during training contains information from $t+1$ and later moves, which the agent can access only if it manages to correctly predict where that information is going to be in its memory space. As a result, when we actually use this policy for playout, the agent's lookups will correspond to places where it would like there to be future information --- which we can then make use of in order to choose the agent's guess. As such, here we are not strictly doing novelty search where we simply look to maximize the variety of the agent's knowledge, but rather we're focusing on the particular question the agent is trying to answer. The network is trained for $150$ epochs on $2\times 10^5$ random games (e.g. where the guesses derive from a uniform random policy) of length 30, using the Adam optimizer with learning rate $1\times 10^{-4}$. 

\begin{figure}
 \includegraphics[width=\columnwidth]{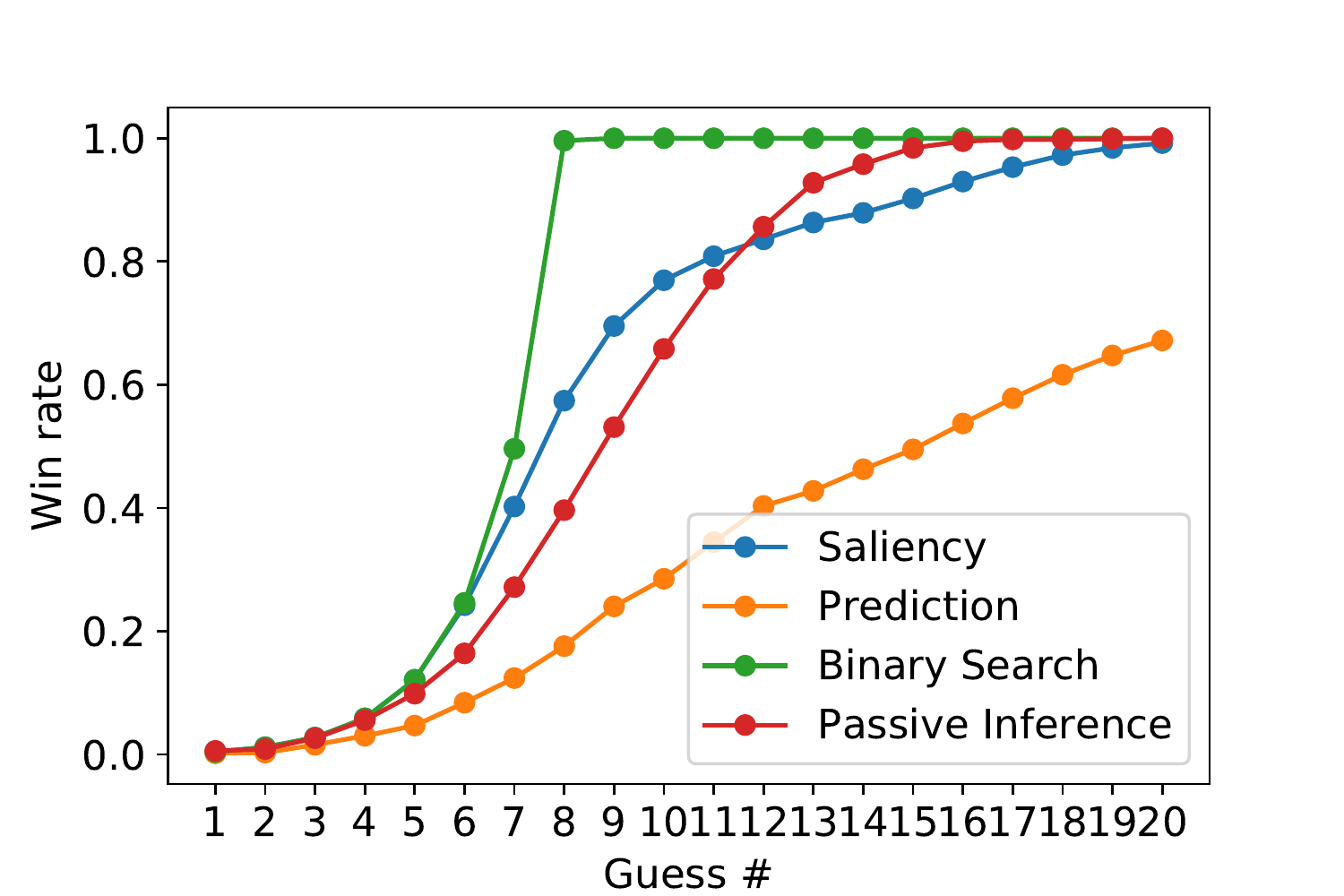}
 \caption{\label{GuessingWin} Success rate by move number of different algorithms in playing the guessing game. A binary search gives the optimal result, shown in green. Passive inference (red curve) in this case corresponds to randomly guessing according to the remaining possible values, and corresponds to the best one can do without some particular strategy for information gain. The saliency profile from the network (blue) outperforms this up to guess 10, showing that the network's saliency successfully performs active inference.}
\end{figure}

Following training, we examine a guessing policy driven by the network's predictions as to the most likely number, and a guessing policy driven by the point of maximum saliency of the third attentional lookup. We find that just guessing according to the saliency pattern alone can implement an active inference strategy which outperforms the best case `passive inference' strategy, of guessing according to the distribution of numbers which are still possible given the evidence up to this point. These results are shown in Fig.~\ref{GuessingWin}. However, the prediction-based strategy significantly underperforms what should theoretically be possible. We have tried a few variant architectures (one using an LSTM to pre-process the guesses so far before the attentional lookup, and another simply reducing all hidden layer sizes to $32$), and it seems that the prediction performance is a result of underfitting. Asking the model to use only three memory lookups in order to summarize up to $20$ guesses seems to be fairly difficult, but at the same time that difficulty causes the network to make good use of the saliency patterns. The LSTM-based architecture performed better at prediction, but its saliency profile was significantly less useful and did not outperform passive inference, while decreasing the hidden layer size damaged both the prediction and saliency performance.

\section{Conclusions}

We examined the possibility of using inference tasks to shape the embedding spaces used by novelty search to differentiate between new and old action policies, using two example systems. In each case, we find evidence that by asking a network to use attentional mechanisms to collect information relevant to its inference task (which can be learned in a differentiable, end-to-end fashion), the sense of `novelty' or `saliency' in the learned space can actually capture some local sensory-motor contigencies and be used to derive a `greedily curious' action policy without the need for any direct reinforcement learning or evolution. These greedy policies do not execute any long-range planning, and as such they cannot always escape uninteresting regions, but can still provide a boost in the rate of exploration of an agent compared to random policy sampling. Therefore, even when the greedy policies are not sufficient on their own to obtain the desired or interesting behaviors, it may be beneficial to use it at the start, and then transition over to policies learned via reinforcement or evolution using the learned embedding space to modify the reward function in the standard novelty search sense.

When the `question' underlying the agent's curiosity is specific in nature, it is possible to use these embedding spaces not to only evaluate the novelty of action policies or outcomes, but also to direct that in a question-dependent way. This means that it is possible to construct an agent which, rather than just being `curious', is `curious about' --- that is to say, that its curiosity is directly tied to its understanding of what is needed to know or answer a specific question. This suggests an interesting interpretation of intrinsic curiosity as a compatible concept to empowerment \citep{klyubin2005empowerment, klyubin2005all}. That is to say, empowerment as an intrinsic motivation maximizes an agent's ability to achieve a diversity of discernable outcomes in the future without knowledge of which specific outcome may correspond to high reward for the agent. Similarly, curiosity independent of specific goals (e.g. `intrinsic' curiosity) could be seen as the maximization of the agent's ability to answer any random question (out of a distribution of potential questions) without prior knowledge of what that question is going to be.

\footnotesize

\bibliographystyle{apalike}
\bibliography{bibliography}

\begin{thebibliography}{}

\bibitem[Achiam and Sastry, 2017]{achiam2017surprise}
Achiam, J. and Sastry, S. (2017).
\newblock Surprise-based intrinsic motivation for deep reinforcement learning.
\newblock {\em arXiv preprint arXiv:1703.01732}.

\bibitem[Ba et~al., 2016]{ba2016layer}
Ba, J.~L., Kiros, J.~R., and Hinton, G.~E. (2016).
\newblock Layer normalization.
\newblock {\em arXiv preprint arXiv:1607.06450}.

\bibitem[Baranes and Oudeyer, 2013]{baranes2013active}
Baranes, A. and Oudeyer, P.-Y. (2013).
\newblock Active learning of inverse models with intrinsically motivated goal
  exploration in robots.
\newblock {\em Robotics and Autonomous Systems}, 61(1):49--73.

\bibitem[Conti et~al., 2017]{conti2017improving}
Conti, E., Madhavan, V., Such, F.~P., Lehman, J., Stanley, K.~O., and Clune, J.
  (2017).
\newblock Improving exploration in evolution strategies for deep reinforcement
  learning via a population of novelty-seeking agents.
\newblock {\em arXiv preprint arXiv:1712.06560}.

\bibitem[de~Abril and Kanai, 2018]{de2018curiosity}
de~Abril, I.~M. and Kanai, R. (2018).
\newblock Curiosity-driven reinforcement learning with homeostatic regulation.
\newblock {\em arXiv preprint arXiv:1801.07440}.

\bibitem[Dickson and Dinner, 2010]{dickson2010enhanced}
Dickson, A. and Dinner, A.~R. (2010).
\newblock Enhanced sampling of nonequilibrium steady states.
\newblock {\em Annual review of physical chemistry}, 61:441--459.

\bibitem[Doncieux and Mouret, 2013]{doncieux2013behavioral}
Doncieux, S. and Mouret, J.-B. (2013).
\newblock Behavioral diversity with multiple behavioral distances.
\newblock In {\em Evolutionary Computation (CEC), 2013 IEEE Congress on}, pages
  1427--1434. IEEE.

\bibitem[Eysenbach et~al., 2018]{eysenbach2018diversity}
Eysenbach, B., Gupta, A., Ibarz, J., and Levine, S. (2018).
\newblock Diversity is all you need: Learning skills without a reward function.
\newblock {\em arXiv preprint arXiv:1802.06070}.

\bibitem[Fiete and Seung, 2007]{fiete2007neural}
Fiete, I.~R. and Seung, H.~S. (2007).
\newblock Neural network models of birdsong production, learning, and coding.
\newblock {\em New Encyclopedia of Neuroscience. Eds. L. Squire, T. Albright,
  F. Bloom, F. Gage, and N. Spitzer. Elsevier}.

\bibitem[Friston et~al., 2015]{friston2015active}
Friston, K., Rigoli, F., Ognibene, D., Mathys, C., Fitzgerald, T., and Pezzulo,
  G. (2015).
\newblock Active inference and epistemic value.
\newblock {\em Cognitive neuroscience}, 6(4):187--214.

\bibitem[Guttenberg et~al., 2017]{guttenberg2017learning}
Guttenberg, N., Biehl, M., and Kanai, R. (2017).
\newblock Learning body-affordances to simplify action spaces.
\newblock {\em arXiv preprint arXiv:1708.04391}.

\bibitem[Houthooft et~al., 2016]{houthooft2016vime}
Houthooft, R., Chen, X., Duan, Y., Schulman, J., De~Turck, F., and Abbeel, P.
  (2016).
\newblock Vime: Variational information maximizing exploration.
\newblock In {\em Advances in Neural Information Processing Systems}, pages
  1109--1117.

\bibitem[Jain and Lindsey, 2018]{jain2018semiparametric}
Jain, M.~S. and Lindsey, J. (2018).
\newblock Semiparametric reinforcement learning.

\bibitem[Kingma and Ba, 2014]{kingma2014adam}
Kingma, D. and Ba, J. (2014).
\newblock Adam: A method for stochastic optimization.
\newblock {\em arXiv preprint arXiv:1412.6980}.

\bibitem[Klyubin et~al., 2005a]{klyubin2005all}
Klyubin, A.~S., Polani, D., and Nehaniv, C.~L. (2005a).
\newblock All else being equal be empowered.
\newblock In {\em European Conference on Artificial Life}, pages 744--753.
  Springer.

\bibitem[Klyubin et~al., 2005b]{klyubin2005empowerment}
Klyubin, A.~S., Polani, D., and Nehaniv, C.~L. (2005b).
\newblock Empowerment: A universal agent-centric measure of control.
\newblock In {\em Evolutionary Computation, 2005. The 2005 IEEE Congress on},
  volume~1, pages 128--135. IEEE.

\bibitem[Lehman and Stanley, 2011]{stanley2011}
Lehman, J. and Stanley, K.~O. (2011).
\newblock Abandoning objectives: Evolution through the search for novelty
  alone.
\newblock {\em Evolutionary Computation}, 19(2):189--223.
\newblock PMID: 20868264.

\bibitem[Meyerson et~al., 2016]{meyerson2016learning}
Meyerson, E., Lehman, J., and Miikkulainen, R. (2016).
\newblock Learning behavior characterizations for novelty search.
\newblock In {\em Proceedings of the Genetic and Evolutionary Computation
  Conference 2016}, pages 149--156. ACM.

\bibitem[Ostrovski et~al., 2017]{ostrovski2017count}
Ostrovski, G., Bellemare, M.~G., Oord, A. v.~d., and Munos, R. (2017).
\newblock Count-based exploration with neural density models.
\newblock {\em arXiv preprint arXiv:1703.01310}.

\bibitem[Paszke et~al., 2017]{paszke2017automatic}
Paszke, A., Gross, S., Chintala, S., Chanan, G., Yang, E., DeVito, Z., Lin, Z.,
  Desmaison, A., Antiga, L., and Lerer, A. (2017).
\newblock Automatic differentiation in pytorch.

\bibitem[Pathak et~al., 2017]{pathak2017curiosity}
Pathak, D., Agrawal, P., Efros, A.~A., and Darrell, T. (2017).
\newblock Curiosity-driven exploration by self-supervised prediction.
\newblock In {\em International Conference on Machine Learning (ICML)}, volume
  2017.

\bibitem[Pritzel et~al., 2017]{pritzel2017neural}
Pritzel, A., Uria, B., Srinivasan, S., Puigdomenech, A., Vinyals, O., Hassabis,
  D., Wierstra, D., and Blundell, C. (2017).
\newblock Neural episodic control.
\newblock {\em arXiv preprint arXiv:1703.01988}.

\bibitem[Pugh et~al., 2016]{pugh2016quality}
Pugh, J.~K., Soros, L.~B., and Stanley, K.~O. (2016).
\newblock Quality diversity: A new frontier for evolutionary computation.
\newblock {\em Frontiers in Robotics and AI}, 3:40.

\bibitem[Saxe et~al., 2013]{saxe2013exact}
Saxe, A.~M., McClelland, J.~L., and Ganguli, S. (2013).
\newblock Exact solutions to the nonlinear dynamics of learning in deep linear
  neural networks.
\newblock {\em arXiv preprint arXiv:1312.6120}.

\bibitem[Torrie and Valleau, 1977]{torrie1977nonphysical}
Torrie, G.~M. and Valleau, J.~P. (1977).
\newblock Nonphysical sampling distributions in monte carlo free-energy
  estimation: Umbrella sampling.
\newblock {\em Journal of Computational Physics}, 23(2):187--199.

\bibitem[Vaswani et~al., 2017]{vaswani2017attention}
Vaswani, A., Shazeer, N., Parmar, N., Uszkoreit, J., Jones, L., Gomez, A.~N.,
  Kaiser, {\L}., and Polosukhin, I. (2017).
\newblock Attention is all you need.
\newblock In {\em Advances in Neural Information Processing Systems}, pages
  6000--6010.

\end{thebibliography}
\end{document}